\documentclass[runningheads]{llncs}
\usepackage[T1]{fontenc}
\usepackage{graphicx, url}
\usepackage{hyperref}
\usepackage{booktabs}
\usepackage{siunitx}
\begin{document}
\title{Standing on the shoulders of giants}
%
%
\author{Lucas F. F. Cardoso\inst{1}\orcidID{0000-0003-3838-3214}\and
José de Sousa R. Filho \inst{2}\orcidID{0000-0002-8836-4188}\and
Vitor C. A. Santos \inst{3}\orcidID{0000-0002-7960-3079}\and
Regiane S. Kawasaki Francês\inst{1}\orcidID{0000-0003-3958-064X}\and
Ronnie C. O. Alves\inst{3}\orcidID{0000-0003-4139-0562}}

\authorrunning{Cardoso, L. et al.}
%
\institute{Federal University of Pará - UFPA, Belém, Brazil\\
\email{lucas.cardoso@icen.ufpa.br, kawasaki@ufpa.br}\and
Federal Institute of Pará - IFPA, Belém, Brazi\\
\email{jose.ribeiro@ifpa.edu.br} \and
Vale Technological Institute - ITV, Belém, Brazil\\
\email{vitor.cirilo.santos@itv.org, ronnie.alves@itv.org}}
\maketitle              
\begin{abstract}
Although fundamental to the advancement of Machine Learning, the classic evaluation metrics extracted from the confusion matrix, such as precision and F1, are limited. Such metrics only offer a quantitative view of the models' performance, without considering the complexity of the data or the quality of the hit. To overcome these limitations, recent research has introduced the use of psychometric metrics such as Item Response Theory (IRT), which allows an assessment at the level of latent characteristics of instances. This work investigates how IRT concepts can enrich a confusion matrix in order to identify which model is the most appropriate among options with similar performance. In the study carried out, IRT does not replace, but complements classical metrics by offering a new layer of evaluation and observation of the fine behavior of models in specific instances. It was also observed that there is 97\% confidence that the score from the IRT has different contributions from 66\% of the classical metrics analyzed.

\keywords{Machine Learning \and Classification \and Item Response Theory.}
\end{abstract}
\section{Introduction}
\label{section1}

Artificial Intelligence (AI) systems are present in all spheres of society. Despite the different types of AI that exist, Machine Learning (ML) techniques are the most commonly used, especially for more objective tasks such as classification problems \cite{lee2020artificial}. The increasing advancement of its applications requires models with increasingly higher success rates and a low chance of error \cite{linardatos2020explainable}.

The binary classification task is a classic ML task, in which the model is challenged with a dataset that has only two classes, normally translated into class 1 (positive) and 0 (negative). This format is used because in binary problems a class tends to be the objective of classification, e.g., in a dataset that has information on patients in a hospital who do or do not have a certain disease, the objective of a classifier would be to identify whether a new patient has the disease (positive) or does not (negative) \cite{sokolova2009systematic}.

With this standard, it is possible to evaluate the model's performance based on its confusion matrix and extract the most classic evaluation metrics used in ML, such as Accuracy, F1 Score and Recall \cite{sokolova2009systematic}. Despite being widely used, classical metrics can only evaluate in a quantitative way, as they only count errors and successes and then evaluate the model over the entire test set, without differentiating the instances that are classified \cite{martinez2019item}.

Recent work aims to overcome this limitation by applying other forms of evaluation within the ML context. An interesting approach that has been gaining ground is the use of concepts from psychometrics to evaluate models. Works such as \cite{cardoso2020decoding,araujo2023quest,de2024explanations} use Item Response Theory (IRT), to more accurately measure the ability of models, due to its main characteristic of being instance-level. With IRT it is possible to evaluate the performance of a model against a specific instance of the test set, in a way that allows the model and data relationship to be further explored. The application of IRT in the context of ML is not to replace classical metrics, but rather to add more information to prior knowledge.

Given the above, this work investigates how IRT can be used to open the box of the classical confusion matrix. The confusion matrix typically delivers its evaluation in a batch of instances format,
therefore, it does not allow observing small fluctuations that the instances cause in the generalization of the models, in their learning. It is observed that IRT is capable of bringing an extra evaluative layer by standing on the shoulders of the confusion matrix and thus reinforcing the assessment of the models' ability, enabling a more assertive decision regarding the comparison between models that eventually perform in a similar way.
A new performance metric was also introduced that evaluates intrinsic aspects of learning, not yet explicitly covered by classical performance metrics.

The remainder of the paper is organized as follows: Section 2 presents the necessary theoretical framework of the article, and explains the concepts of using IRT applied to ML; Section 3 deals with the methodology used to carry out the experiments proposed in the article; Section 4 presents the results obtained by the experiments and discusses what was observed; Section 5 concludes the article and presents future work.

\section{Background}
\label{section2}

\subsection{Item Response Theory}
\label{section2.2}

The objective of this section is to explain the psychometric concepts of IRT, how it works and how it can be used in Machine Learning. As already mentioned in section \ref{section1}, IRT emerged within psychometrics as a new way of evaluating individuals' performance on the same test. Just like in classic ML metrics, the classic way of evaluating people is done through a test where performance is measured based on the individual's mistakes or successes in the test, so that 1 is added for each correct answer and 0 for each error. Although it is still widely used, this format is not capable of measuring an individual's real ability, as it does not consider the complexity of the different items that exist in the test \cite{baker2001basics}.

The IRT then emerges as a new assessment approach that is centered on the items and not on the test itself. Its objective is to calculate the probability of success of a given respondent $j$ on a specific item $i$ considering the ability $\theta$ that this individual has and the parameters that describe item $i$. There are 3 main parameters: ($a_{i}$) discrimination, whose value represents how well a given item is able to distinguish respondents with high ability from respondents with low ability, the higher its value, the more discriminating the item is; ($b_{i}$) difficulty, restates how difficult an item is to answer correctly, the higher its value, the more difficult an item is; ($c_{i}$) guessing, the parameter that indicates the minimum chance that an item has of being answered correctly or the chance of a low-skill respondent getting the item right at random \cite{cai2016item}.

IRT has different logistic models that are used to calculate the probability of a correct answer, determined by the number of item parameters they use. The logistic model that uses the three item parameters is called the Three-Parameter Logistic model (3PL) for dichotomous items, i.e., items where only whether it was answered correctly or not is considered. The 3PL logistic model is defined by Equation \ref{eq:3pl}, where $P(U_{ij} = 1\vert\theta_{j})$ is the probability of the respondent $\theta_{j}$ correctly answering the dichotomous item $U_{ij}$ \cite{baker2001basics,cai2016item}.

\begin{equation} \label{eq:3pl}
    P(U_{ij} = 1\vert\theta_{j}) = c_{i} + (1 - c_{i})\frac{1}{1+ e^{-a_{i}(\theta_{j}-b_{i})}}
\end{equation}

Despite being item-centered, IRT is also capable of generating a final score on an individual's performance. To do this, you can use the concept of True Score \cite{lord1984comparison}, which generates a score $TrueS_{j}$ for each respondent based on the sum of the probabilities of a correct answer $P(U_{ij} = 1\vert \theta_{j})$ calculated for each test item, the True Score is defined by Equation \ref{eq:true_score}. 

\begin{equation} \label{eq:true_score}
    TrueS_{j} = \sum_{i=1}^{N} P(U_{ij} = 1\vert\theta_{j})
\end{equation}

\subsubsection{Item Characteristic Curve.}
\label{section2.2.1}

\begin{figure}[!ht]
\centering
\includegraphics[width=0.8\textwidth]{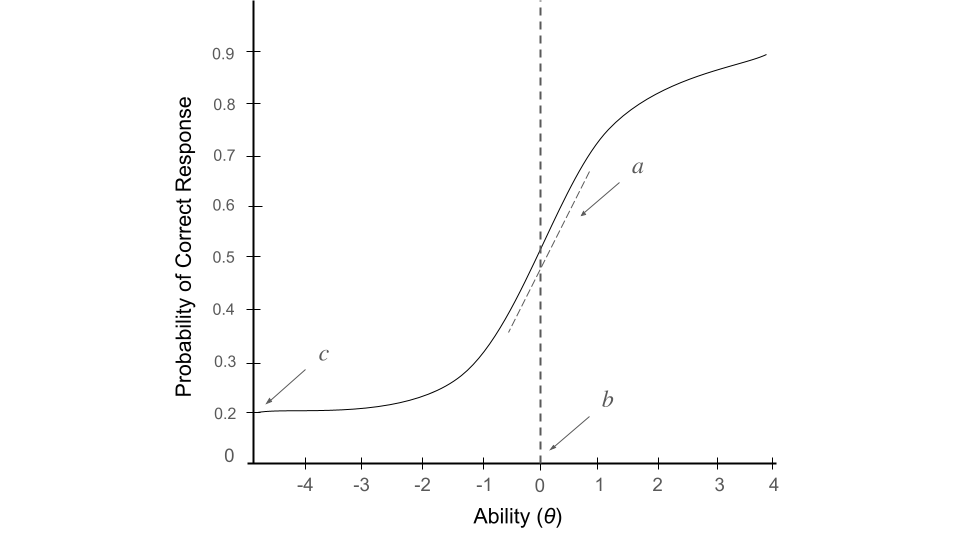}
\caption{Example of Item Characteristic Curve.} \label{fig:icc}
\end{figure}

One of the main products of the IRT is the Item Characteristic Curve (ICC), which allows analyzing the relationship between item parameters and the probability of a correct answer given the variation in ability. Figure \ref{fig:icc} presents an example of the ICC. It is possible to visualize the behavior of the probability of a correct answer increasing as the ability $\theta$ also increases. In the ICC the guessing parameter $c$ determines the lowest probability of a correct answer that the curve will have, while the discrimination $a$ determines the slope of the curve, so that the higher the value of $a$ the steeper the curve becomes. The difficulty $b$ and the ability $\theta$ are calculated on the same scale, so that if $\theta=b$ the chance of a correct answer is equal to 50\%. The difficulty determines the point on the curve where this happens \cite{baker2001basics}.

\section{Methodology}
\label{section3}

In this section, the methodology used to carry out the experiments with the application of IRT in ML will be presented. As mentioned in Section \ref{section1}, this work aims to explore how IRT can be useful for evaluating ML models considering the confusion matrix obtained in a classical classification problem\footnote{Link to the source code of the methodology: \url{https://github.com/LucasFerraroCardoso/IRT_Confusion_Matrix}}.

Although its use is aimed at evaluating individuals, as explained in section \ref{section2.2}, a simple analogy is enough to apply IRT in the context of ML when considering respondents as models while items are test instances. To calculate the probability of a correct answer, the 3PL logistic model was used, due to the fact that it considers the probability of a random hit, as it is possible for the models to get the instances right by guessing. To estimate the item parameters of the dataset instances and the ability of the models, the Birnbaum method \cite{martinez2016making} was used, where the item parameters are first estimated by considering only the responses of all models and subsequently the ability is estimated from the item parameters obtained and the response vector of each model.

\begin{figure}[!ht]
\includegraphics[width=1\textwidth]{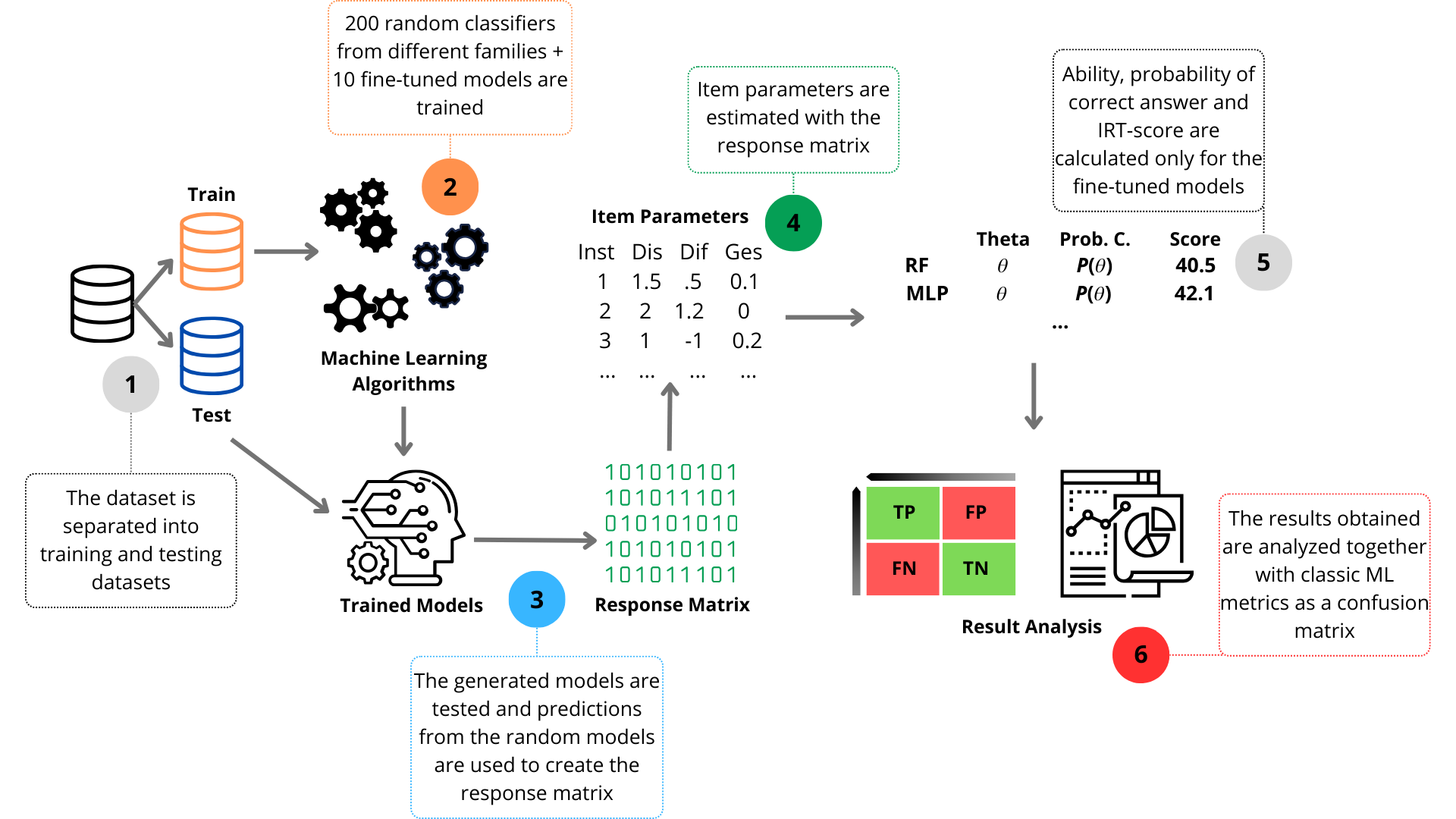}
\caption{Flowchart of the methodology.} \label{fig1}
\end{figure}

As a case study, the Heart-Statlog\footnote{Heart-Statlog Link: \url{https://www.openml.org/search?type=data&status=active&id=53}} dataset was chosen. The Heart-Statlog is a heart disease dataset, which is composed of 270 instances with 13 features. Figure \ref{fig1} illustrates the methodology used to calculate the IRT estimators, which is separated into 6 main steps.

\begin{enumerate}
    \item The Heart-Satolog dataset is separated into training and testing. The necessary pre-processing of the data is also carried out at this stage, such as coding categorical features.
    \item Next, 200 random models are generated from different classification algorithms that will be trained with the training dataset. At this stage, 10 fine-tuned models are also trained, whose performance will be evaluated later.
    \item The trained random models must then classify the instances of the test dataset to generate the response matrix from their predictions. The fitted models also classify the test dataset. The response matrix is a matrix where the columns are the instances and the rows are the dichotomous response vectors for each instance, with 1 being correct and 0 being incorrect.
    \item The response matrix that was obtained is used to estimate the item parameters of the test dataset.
    \item With the item parameters calculated, it is possible to estimate the ability of the fine-tuned models and then calculate the probability of a correct answer and the IRT scores.
    \item The results obtained are then divided according to the confusion matrix to analyze the performance of the models.
\end{enumerate}

To estimate accurate values of item parameters, 200 random classifiers from 10 different learning algorithms were used (Decision Tree (DT), Random Forest (RF), Ada Boost (ADA), Gradient Boosting (GB), Bagging (BAG), Multilayer Perceptron (MLP), k-Nearest Neighbors (KNN), Support Vector Machine (SVM), Linear Support Vector Machine (LSVM) and Linear Discriminant Analysis (LDA). To achieve this, 20 models were generated with randomly chosen hyperparameters for each of the 10 algorithms used. This way, it is possible to generate a diversity of answers, as if there were different students of different levels answering a test. The Scikit-Learn library \cite{pedregosa2011scikit} was used to generate all the classification algorithms.

For performance analysis, 10 models were generated, one for each algorithm used. These 10 models are trained and hyper-parameterized using the grid search strategy with cross-validation provided by Scikit Learn, with 5-folds for cross-validation. The R package Ltm \cite{rizopoulos2006ltm} was used to estimate the IRT item parameters from the response matrix. Subsequently, the ability of the fitted models was estimated using the Catsim \cite{meneghetti2017application} Python package, this same package was used to calculate the probability of a correct answer and the ICC's of the test instances.

Although IRT already has the True-Score, presented in section \ref{section2.2}, it is understood that it is important to also consider the error for a fairer evaluation of the models. For this, the concept of Total-Score, presented in \cite{de2024explanations}, was used, where the probability of error is subtracted from the score for instances that were incorrectly classified. The Total-Score is defined by the Equation \ref{eq:total_score}, where $i'$ corresponds to items that were answered correctly while $i''$ corresponds to items answered incorrectly.

\begin{equation} \label{eq:total_score}
    TotalS_{j} = \sum_{i=1}^{i'} P(U_{ij} = 1\vert\theta_{j}) - \sum_{i=1}^{i''} 1 - P(U_{ij} = 1\vert\theta_{j})
\end{equation}

\section{Results and Discussion}
\label{section4}

In this section, the results obtained from the experiments carried out in this research are presented and discussed\footnote{All results generated are available at the link: \url{https://github.com/LucasFerraroCardoso/IRT_Confusion_Matrix/tree/main/Results}}. As presented in section \ref{section3}, the Heart-Statlog dataset was chosen as a case study because it is a known and easy-to-understand dataset. The Heart-Statlog is a dataset that contains data from patients who may or may not have heart disease. It has 270 instances with 13 patient diagnostic features, with 120 instances of patients who have heart disease (positive class) and 150 instances of patients who do not have heart disease (negative class), so the dataset has a balance of 55.55\% for the negative class (majority) and 44.45\% for the positive class (minority). In this work, the division 70\% was used for training and 30\% for testing.

\begin{figure}[!ht]
\centering
\includegraphics[width=0.65\textwidth]{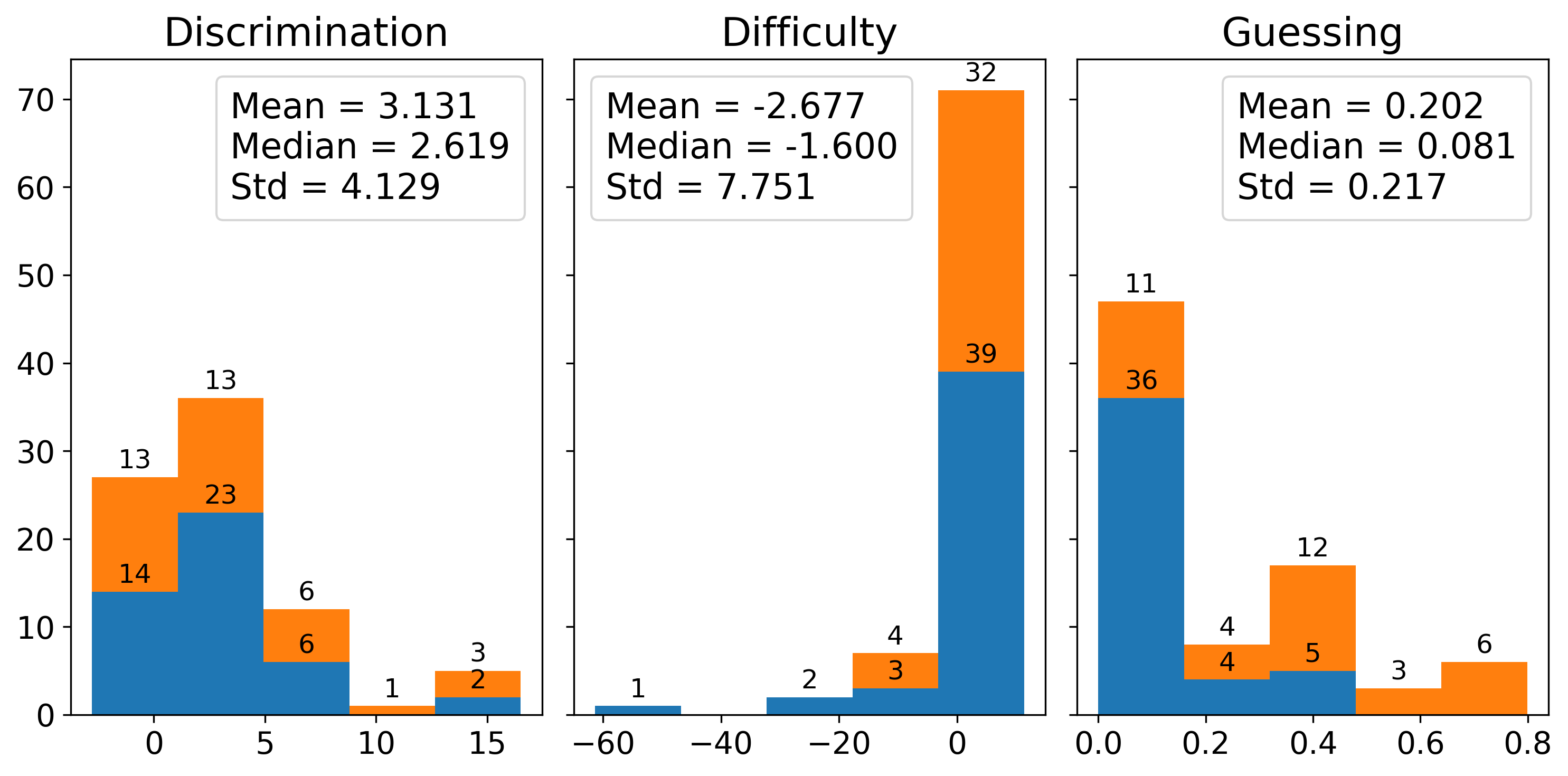}
\caption{Heart-Statlog item parameter histograms. The orange bars are the instances of the minority class while the blue ones are the majority class.} \label{fig2}
\end{figure}

To understand the performance of models, it is important to first understand the complexity of the dataset. Figure \ref{fig2} generally shows the item parameters estimated for the Heart-Statlog, in which the instances were organized into a histogram for each item parameter separating the majority and minority class instances. According to IRT concepts, the dataset can be considered not very difficult due to the negative average of difficulty with -2.67 and suitable for comparing respondents due to the positive average of discrimination with 3.13.

However, despite the positive average discrimination, it is important to highlight that the dataset presents around 24.7\% of instances with negative discrimination. While instances with high discrimination values mean that they are good at certifying the models' ability, instances with negative values mean that they are not well formed or even defined, so if a model hits many instances with these characteristics it means that this model may not be reliable.

It is also noted that there is a balance between the frequency of majority and minority instances present in the 5 bins for the difficulty and discrimination histograms, this was already expected given the balance of 55\%-45\% between the classes. For the guessing parameter, the minority class presented the highest frequency of instances with a high probability of random success, despite balancing. Previous research has already highlighted that models tend to hit instances of the minority class by chance more frequently \cite{cardoso2022explanation}.

\begin{figure}[!ht]
\centering
\includegraphics[width=0.6\textwidth]{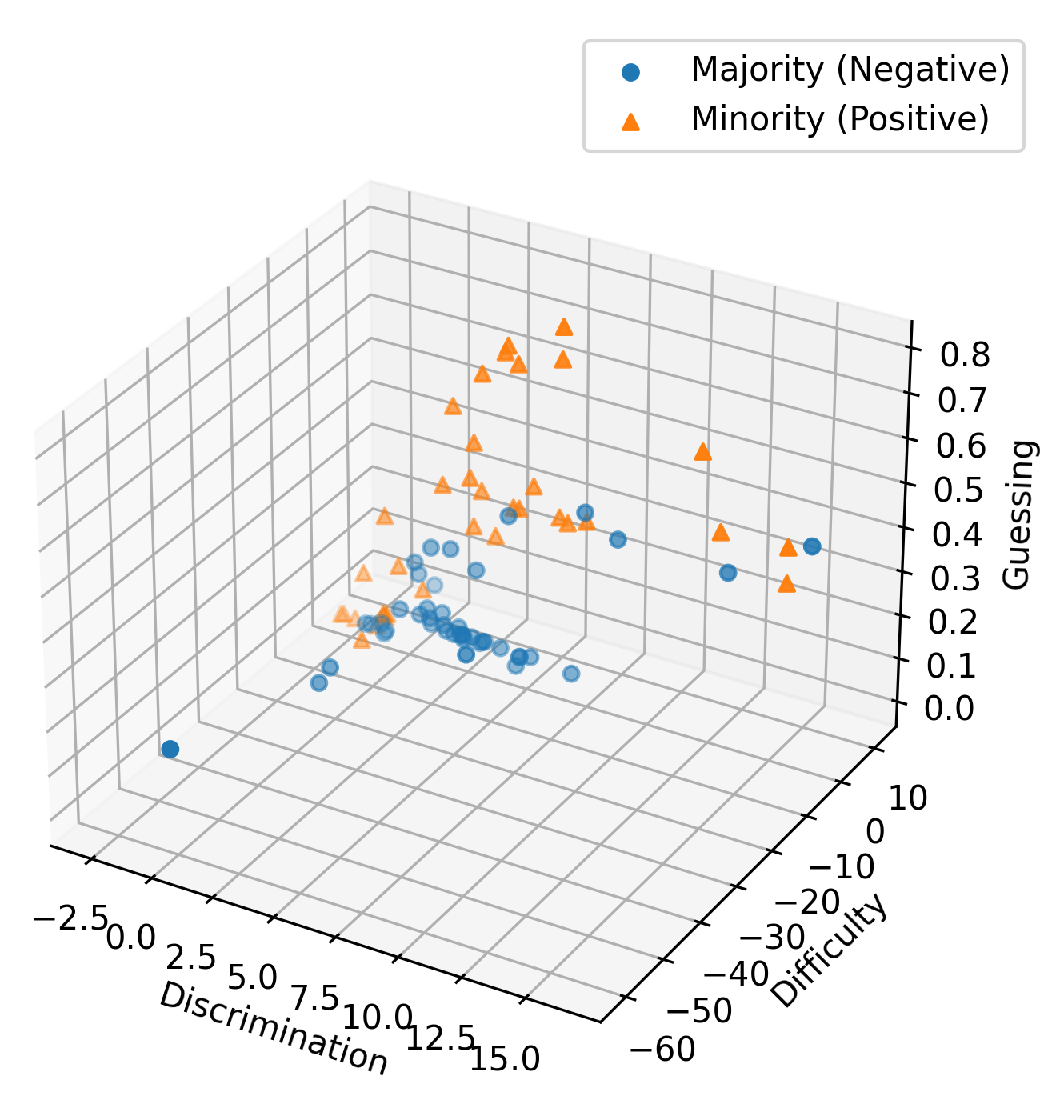}
\caption{Heart-Statlog instances arranged by item parameters.} \label{fig:3d}
\end{figure}

This distribution of instances by the guessing parameter becomes even more evident when analyzing Figure \ref{fig:3d}. Although the largest number of instances of the majority class have 0 guessing, which means that chance does not tend to influence the correct answer, there are still 9 instances (11\%) that have a greater chance of being classified by guessing. From the figure it is also evident that some instances of the minority class have difficulty outliers, where there are three instances that are considered much easier than the others because they have very negative values. It is understood that in a dataset it is expected that there will be determining examples of a class distribution.

Once you understand the behavior of the data, you can then look at the classifiers. Table \ref{tab1} presents the results of classic ML metrics for the 10 fine-tuned models with their respective ranking positions. When observing only the classic metrics, the GB model presents the highest score achieved in 5 of the 6 metrics considered. If the objective of this experiment was to choose the best model, with a brief analysis it is possible to indicate that the GB would be ideal. However, GB only has the 6th best score if only the recall is considered and in this case this is very pertinent information. Simply put, the recall metric represents how well a model was able to correctly classify True Positives instances, and in the case of a context-sensitive dataset that aims to correctly identify patients who have heart disease, the recall metric has great relevance. If only recall is considered, the model that presents the best performance is LSVM.

\begin{table}[!ht]
\caption{Scores obtained by each model for classic ML metrics.}
    \label{tab1}
    \centering
    \begin{tabular}{@{}|l|c|c|c|c|c|c|@{}}
    \hline
&  \textbf{accuracy} & \textbf{F1 score} & \textbf{precision} & \textbf{recall} & \textbf{auc score} & \textbf{specificity} \\
\hline
RF   & 0.802 (2)   & 0.778 (2)   & 0.778 (5)   & 0.778 (2)   & 0.800 (2)   & 0.822 (5)   \\
GB   & \textbf{0.827 (1)}   & \textbf{0.781 (1)}   & \textbf{0.893 (1)}   & 0.694 (6)   & \textbf{0.814 (1)}   & \textbf{0.933 (1)}   \\
BAG   & 0.790 (5)   & 0.761 (5)   & 0.771 (6)   & 0.750 (3)   & 0.786 (6)   & 0.822 (5)   \\
ADA   & 0.765 (8)   & 0.725 (8)   & 0.758 (7)   & 0.694 (6)   & 0.758 (8)   & 0.822 (5)   \\
KNN   & 0.617 (10)   & 0.523 (10)   & 0.586 (10)   & 0.472 (10)   & 0.603 (10)   & 0.733 (10)   \\
DT   & 0.728 (9)   & 0.686 (9)   & 0.706 (9)   & 0.667 (9)   & 0.722 (9)   & 0.778 (8)   \\
SVM   & 0.790 (5)   & 0.754 (7)   & 0.788 (4)   & 0.722 (5)   & 0.783 (7)   & 0.844 (3)   \\
MLP   & 0.802 (2)   & 0.771 (4)   & 0.794 (3)   & 0.750 (3)   & 0.797 (3)   & 0.844 (3)   \\
LDA   & 0.802 (2)  & 0.758 (6)   & 0.833 (2)   & 0.694 (6)   & 0.792 (4)   & 0.889 (2)   \\
LSVM   & 0.790 (5)  & 0.773 (3)   & 0.744 (8)   & \textbf{0.806 (1)}   & 0.792 (4)   & 0.778 (8)   \\
\hline    
    \end{tabular}
\end{table}


Although LSVM has the best recall score, the model only has the 5th best average performance among all classic metrics. RF has the second best average performance, behind only GB, and has the second best recall score, a condition that would qualify it as a candidate to be chosen as the best model. The precision and recall metrics tend to have inverse performances, i.e. to increase one the other tends to decrease, and this occurs in this example where the RF only presents the 5th precision score. If you also consider the high variation in the ranking as an elimination criterion, it is the MLP that has the 3rd best performance on average and also has the lowest variation in position among the rankings, with high values in both precision and recall. \textit{In this scenario, what would be the best model to choose?} \textit{How else can models be evaluated appropriately?}

To try to answer these questions, the performance of the models will be analyzed from the perspective of IRT. The first step was to calculate the IRT scores proposed in section \ref{section3}. For the True Score, which only considers the respondents' probability of being correct, it was the BAG model that appears first, and it is important to highlight that the difference between True Scores is only 0.0002 for the second (GB). For the Total Score that considers whether there was an error or a correct answer, GB appears again with the highest score, this helps to reaffirm its position as the best model in this scenario.

\begin{table}[!ht]
\caption{True and Total Scores for the models.}
    \label{tab3}
    \centering
    \begin{tabular}{@{}|l|c|c|c|c|@{}}
    \hline
&  \textbf{True Score} & \textbf{TrueS Rank} &  \textbf{Total Score} & \textbf{TotalS Rank} \\
\hline
RF      & 0.7849      & 3      & 0.5874      & 2      \\
GB      & 0.7851      & 2      & \textbf{0.6122}      & \textbf{1}      \\
BAG      & \textbf{0.7853}      & \textbf{1}      & 0.5754      & 4      \\
ADA      & 0.7603      & 9      & 0.5258      & 8      \\
KNN      & 0.6692      & 10      & 0.2865      & 10      \\
DT      & 0.7659      & 6      & 0.4943      & 9      \\
SVM      & 0.7790      & 5      & 0.5691      & 5      \\
MLP      & 0.7651      & 7      & 0.5676      & 6      \\
LDA      & 0.7840      & 4      & 0.5865      & 3      \\
LSVM      & 0.7651      & 7      & 0.5552      & 7      \\
\hline    
    \end{tabular}
\end{table}

\begin{figure}[!ht]
\centering
\includegraphics[width=0.85\textwidth]{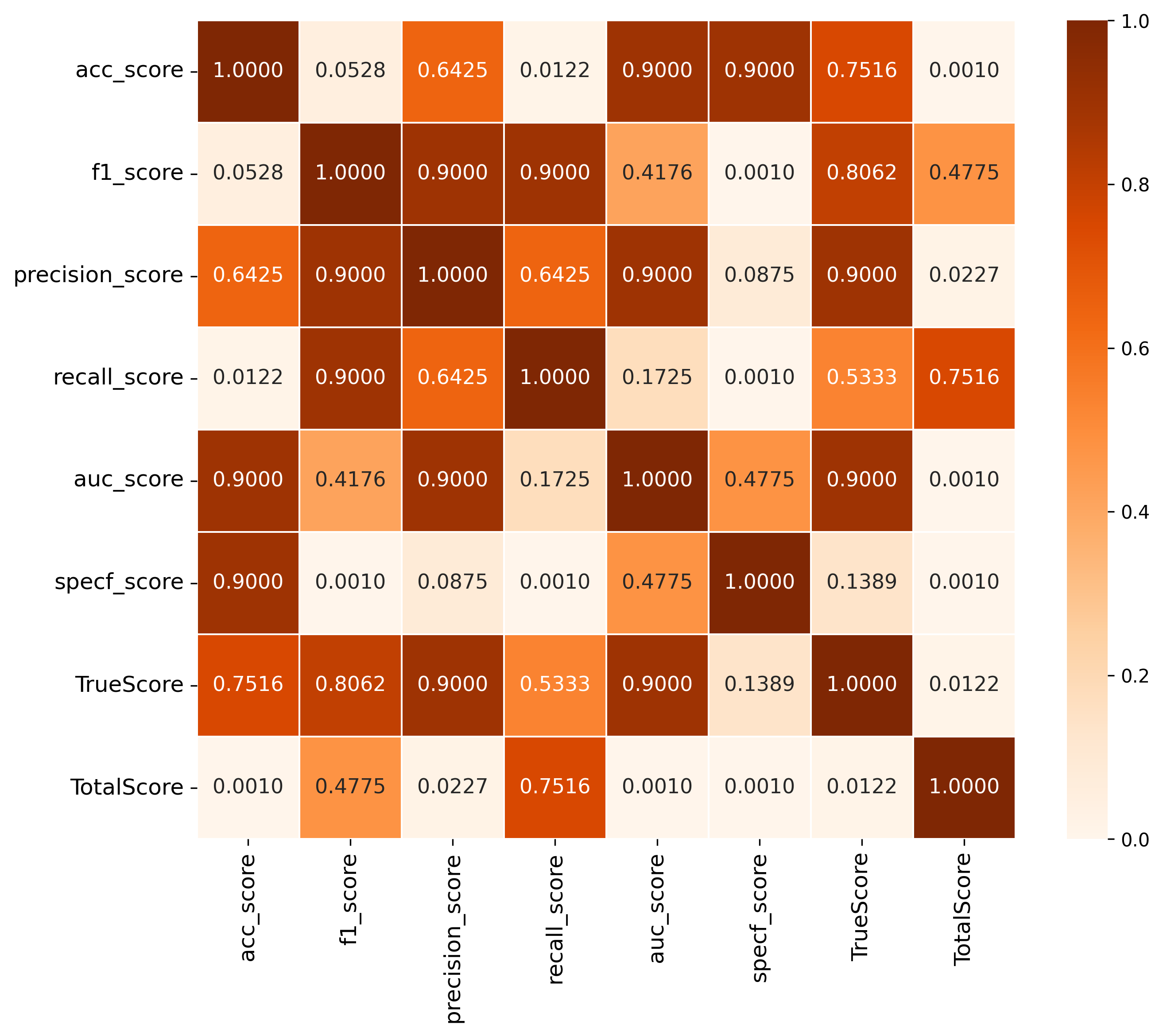}
\caption{Nemenyi test heatmap for evaluation metrics.} \label{fig:heatmap}
\end{figure}

To assess whether there is a gain in using IRT scores in conjunction with classical metrics, the Friedman \cite{pereira2015overview} and Nemenyi \cite{nemenyi1962distribution} tests were performed on the results of all metrics to confirm whether there are different distributions. The Friedman test is a non-parametric test used to check whether there are different distributions among sampled data. Its result is a p-value, which if it is less than a certain limit, then it is considered that there are different distributions in the data. The Nemenyi test is complementary and is used to check which distributions are different when performing pairwise comparisons. The Friedman test resulted in a \textit{p-value} of approximately \num{1.9e-09}, when considering a \textit{p-value} threshold of \num{0.05} it can be stated that there are different distributions between scores. Next, a Nemenyi test was performed to verify which evaluation metrics are different.

Figure \ref{fig:heatmap} presents the result of the Nemenyi test, where it is possible to see that the \textit{p-value} values of the True Score do not show any difference from the classic metrics. On the other hand, interesting results are observed for the Total Score, where the f1 and recall metrics are the only ones that do not present a \textit{p-value} smaller than \num{0.05}. Thus, applying the confidence measure, there is 52.25\% confidence that f1 is statistically different from the Total Score, while there is 24.84\% confidence that the recall is statistically different, for the other metrics the lowest confidence is 97.73\%. Such results point to the Total Score as a metric capable of giving visibility to new nuances in the models’ performance.

To confirm the result obtained, the instances were then opened using the concepts of the IRT Item Characteristic Curves. An ideal confusion matrix should present values only on its main diagonal. With this in mind, the ICC's of each instance were separated according to the models' confusion matrix, called the Item Characteristic Confusion Matrix Curve (ICCMC).

\begin{figure}[!ht]
\centering
\includegraphics[width=0.9\textwidth]{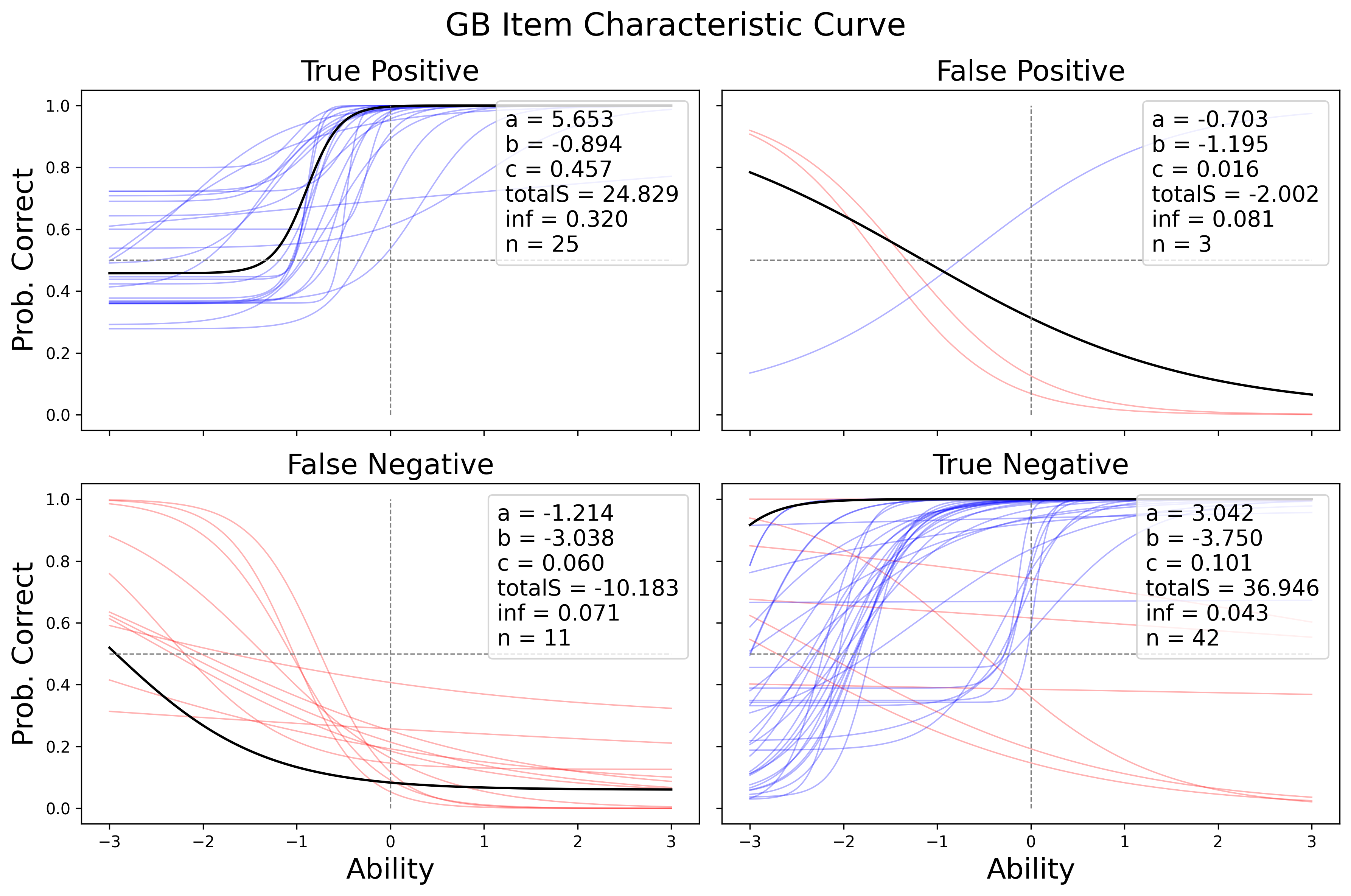}
\caption{ICCMC of the GB model, where $a$ is the average discrimination value, $b$ is the average difficulty, $c$ is the average guessing, $totalS$ is the Total Score value, $inf$ is the average information of the instances and $n$ is the number of instances. The blue lines represent instances with positive discrimination, the red lines are instances with negative discrimination, and the black line is the average ICC of the instances.} \label{fig3}
\end{figure}

Figure \ref{fig3} shows the ICCMC of the GB model, in which it is possible to observe that the model performs well for True Positives and True Negatives instances. Therefore, all TP instances have high discrimination values and are considered good instances to attest to the model's ability. Furthermore, only one well-formed instance appears among the FP, it is expected that only instances with negative discrimination appear between the FP and FN, as can be observed in the set of FN instances of the GB model. The point of greatest attention is the TN set, where 7 (16.67\%) are instances that are not suitable for evaluating the model's performance (red lines). This can also be observed by the average information value of the instances, where the set of TPs presents a high average information value with 0.320, while the TNs present a low average information value with 0.043.

\begin{figure}[!ht]
\centering
\includegraphics[width=0.9\textwidth]{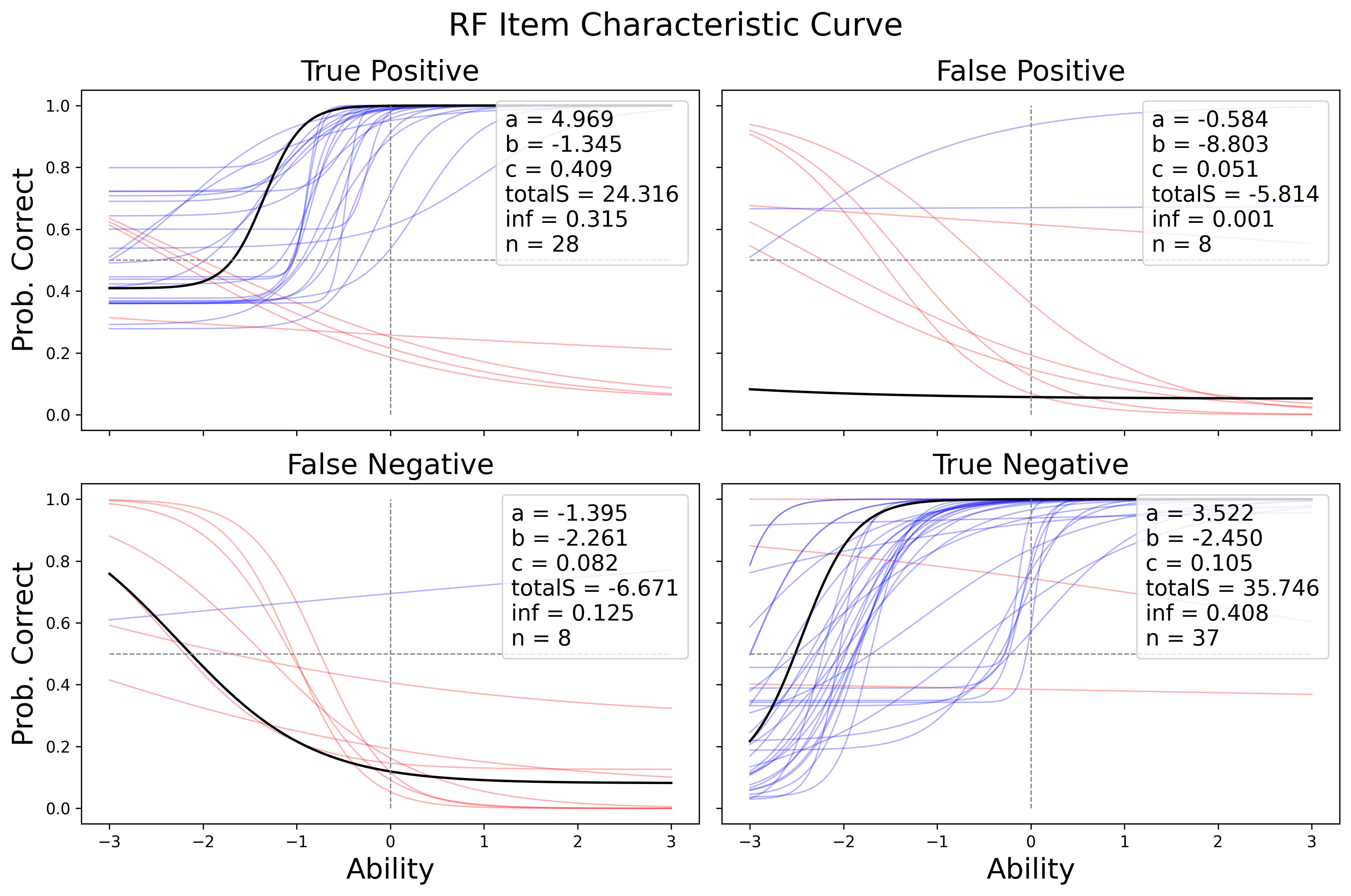}
\caption{ICCMC of the RF model.} \label{fig4}
\end{figure}

In direct comparison with the ICCMC of the RF, it is highlighted that the RF has a lower performance compared to the GB (see Figure \ref{fig4}). This can be confirmed by analyzing the sets of TP and TN instances, where the GB Total Score was higher in both cases. It is also noted that, although RF hits more TP instances (28) than GB (25), 4 instances classified by RF present negative discrimination. In addition, 3 well-formed instances appear in the FP and FN sets. Although the Recall of RF is greater than GB, the set of TP instances of GB is composed only of well-formed items while RF presents instances that are not suitable for evaluating the model.

Another interesting detail is the high guessing value for the instances of the TP set seen in all ICCMC presented, as previously observed in Figure \ref{fig3}, the minority class has a high chance of getting it right by chance. This raises the question of whether the models have actually learned how to classify this class or are just getting lucky. A possible future work would be to investigate more incisively the impact of the guessing parameter and whether these instances are really well formed.

\begin{figure}[!ht]
\centering
\includegraphics[width=0.9\textwidth]{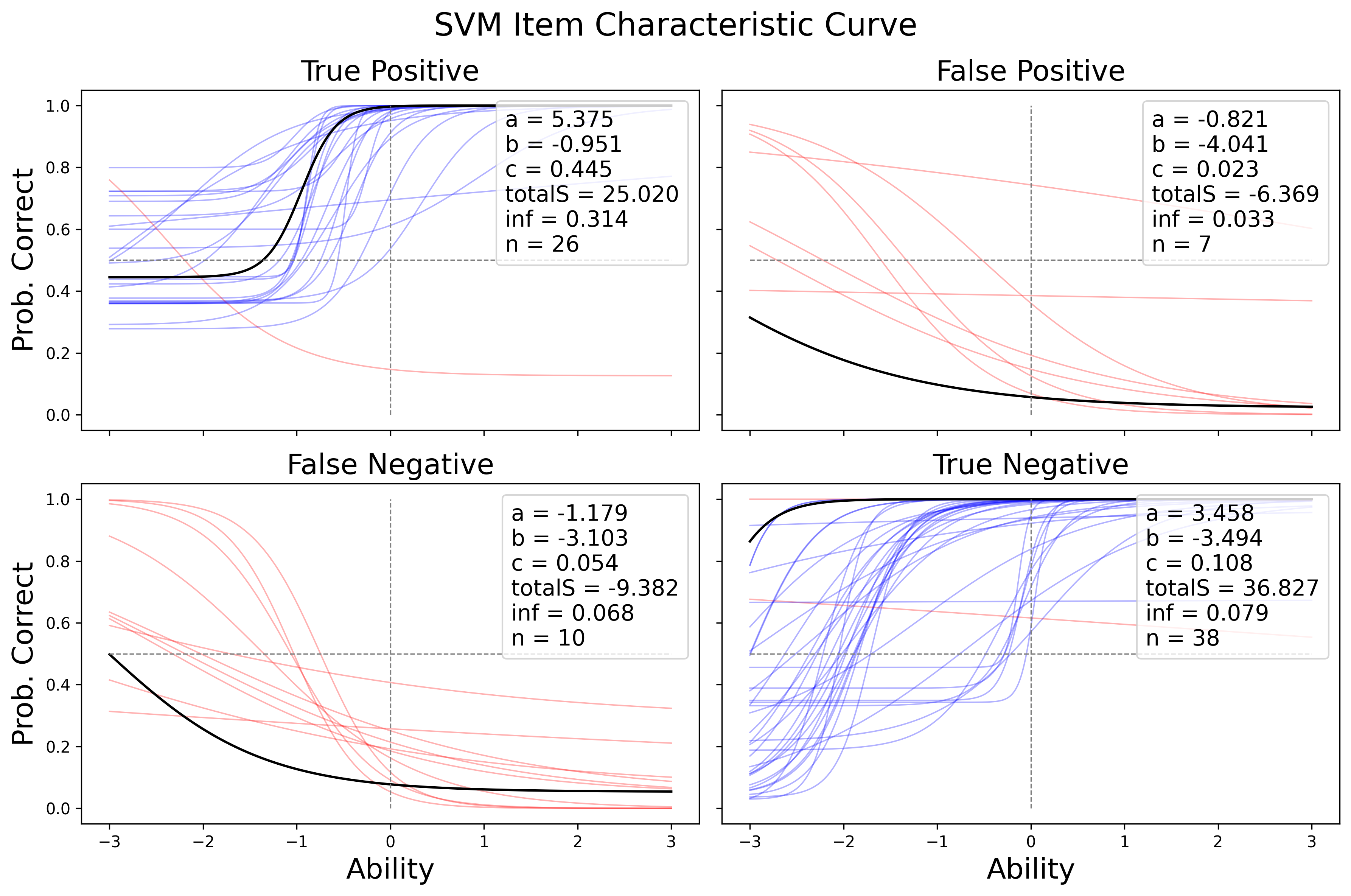}
\caption{ICCMC of the SVM model.} \label{fig5}
\end{figure}

Other models with interesting results are SVM and LDA, where the best configuration of ICC curves of the instances arranged in the confusion matrix sets is noted. As can be seen in Figure \ref{fig5} for the SVM there is only one instance considered not suitable for evaluation in the TP set and only instances with negative discrimination in the FP and FN sets as expected in an ideal scenario. For the TN set, there are only 2 malformed instances, but it is interesting that among them there is an instance that is repeated as TN in all models. For LDA, the TP set is made up of instances suitable for evaluation, while there are three instances with negative discrimination (red lines) in the TN set.

In the work \cite{smith2014reducing} the authors indicate that not all instances of a dataset are equally good for learning, it is common knowledge that models need quality data to be trained. If the same thinking is applied to the evaluation, where instances with negative discrimination are removed from the analysis because they are considered bad by the IRT, the result would be that LDA and SVM would present the best performance in all classical ML metrics (see Table \ref{tab4}). Furthermore, when analyzing the ICCs within the confusion matrix, LDA would be preferable because it presents the most reliable TP set for the Heart-Statlog scenario. Furthermore, another point that makes LDA preferable is its explanatory nature \cite{arrieta2020explainable}. In a context-sensitive scenario such as the one analyzed, a simpler model to explain its predictions is also more suitable.

Which model is ideal for a given dataset will depend on the context in which it is inserted and determining the ideal model should only be done after a more thorough analysis of performance considering the complexity of the data. A deeper analysis of the relationship between instances at the feature level would be interesting to further explore the use of ICC's in the confusion matrix.

\begin{table}[!ht]
\caption{Ranking of scores obtained for classic metrics without negative discrimination.}
    \label{tab4}
    \centering
    \begin{tabular}{@{}|l|c|c|c|c|c|c|@{}}
    \hline
&  \textbf{accuracy} & \textbf{F1 score} & \textbf{precision} & \textbf{recall} & \textbf{auc score} & \textbf{specificity} \\
\hline
RF                            & 0.951               & 0.941              & 0.923                     & 0.96                   & 0.952               & 0.944                 \\
GB                            & 0.984               & 0.980              & 0.962                     & 1.00                   & 0.986               & 0.972                 \\
BAG                           & 0.918               & 0.902              & 0.885                     & 0.92                   & 0.918               & 0.917                 \\
ADA                           & 0.918               & 0.902              & 0.885                     & 0.92                   & 0.918               & 0.917                 \\
KNN                           & 0.738               & 0.667              & 0.696                     & 0.64                   & 0.723               & 0.806                 \\
DT                            & 0.869               & 0.846              & 0.815                     & 0.88                   & 0.871               & 0.861                 \\
SVM                           & \textbf{1.000}               & \textbf{1.000}              & \textbf{1.000}                     & \textbf{1.00}                   & \textbf{1.000}               & \textbf{1.000}                 \\
MLP                           & 0.967               & 0.962              & 0.926                     & 1.00                   & 0.972               & 0.944                 \\
LDA                           & \textbf{1.000}               & \textbf{1.000}              & \textbf{1.000}                     & \textbf{1.00}                   & \textbf{1.000}               & \textbf{1.000}                 \\
LSVM                          & 0.934               & 0.926              & 0.862                     & 1.00                   & 0.944               & 0.889      \\
\hline    
    \end{tabular}
\end{table}

\section{Final Considerations}
\label{section5}

The empirical evaluation of ML models remains the most common way to analyze the performance of a classifier on a dataset. This work was developed with the aim of deepening the classic form of evaluation considering the performance of models at the instance level using IRT. A case study was presented where the aim was to choose the best model for a given heart disease classification problem with the Heart-Satlog dataset. IRT allowed a deeper understanding of the relationship between model and data, by combining the traditional way of evaluating the confusion matrix with the psychometric concepts of IRT, represented by the idea of the Item Characteristic Confusion Matrix Curve (ICCMC). Furthermore, the direct relationship between IRT scores and classic metrics was explored and it was shown that the Total Score has statistically significant differences in 66\% of the evaluated metrics, with a confidence of 97\%. An experiment was carried out with 10 models from 10 different classification algorithms, while the classic metrics pointed to GB as the most suitable model, the IRT estimators allowed us to understand the data and how the models behaved in each specific instance. In a scenario where the entire test dataset is used for evaluation, IRT allowed us to point out with greater confidence that the GB would be the most suitable, but in another scenario with a filter of the evaluated instances, the LDA model would be the most suitable. If it was possible to see further in models' abilities it is by standing on the shoulders of confusion matrix.

As future work, it is interesting to explore more datasets of different types and contexts, and evaluate whether it would be possible to derive a new metric that would allow evaluating models directly considering the complexity of the data in relation to its features. Another aspect to be observed is the limitations of the methodology used. It is noted that the results obtained by the IRT are conditioned by the population of respondents and items selected for evaluation. Thus, the results may be different if another set of models and items is selected. The IRT itself already has tools to perform the evaluation considering different populations; it is interesting to explore this in future work.

\begin{credits}

\subsubsection{\discintname}
The authors have no competing interests to declare that are relevant to the content of this article.
\end{credits}
%
%
%
\bibliographystyle{splncs04}
\bibliography{mybibliography}
\end{document}